\documentclass[runningheads]{llncs}
\usepackage{graphicx}

\usepackage{tikz}
\usepackage{comment}
\usepackage{amsmath}
\usepackage{amssymb} %
\usepackage{color}
\usepackage{xspace}
\usepackage{bm}
\usepackage{dsfont}
\usepackage{booktabs}
\usepackage{multirow}
\usepackage{algorithm,algpseudocode}

\begin{document}

\pagestyle{headings}
\mainmatter

\title{PointInst3D: Segmenting 3D Instances by Points} %
 
\author{Tong He$^{1,3}$, Wei Yin$^{1,4}$, Chunhua Shen$^2$, Anton van den Hengel$^1$}
\institute{1. The University of Adelaide, 2. Zhejiang University, 3. Shanghai AI Lab, 4. DJI}

\def\etal{{\it et al.}\xspace}
\def\eg{{\it e.g.}\xspace}
\def\ie{{\it i.e.}\xspace}

\maketitle

\begin{abstract}

The current state-of-the-art methods in 3D instance segmentation typically involve a clustering step, despite the tendency towards heuristics, greedy algorithms, and a lack of robustness to the changes in data statistics.  
In contrast, we propose a fully-convolutional 3D point cloud instance segmentation method that works in a per-point prediction fashion. In doing so it avoids the challenges that clustering-based methods face: introducing dependencies among different tasks of the model. We find the key to its success is assigning a suitable target to each sampled point. Instead of the commonly used static or distance-based assignment strategies, we propose to use an Optimal Transport approach to optimally assign target masks to the sampled points according to the dynamic matching costs.
Our approach achieves promising results on both ScanNet and S3DIS benchmarks. 
The proposed approach removes inter-task dependencies and thus represents a simpler and more flexible 3D instance segmentation framework than other competing methods, while achieving improved segmentation accuracy.
\end{abstract}

\section{Introduction}

3D instance segmentation describes the problem of identifying a set of instances that explain the locations of a set of sampled 3D points. It is an important step in a host of 3D scene-understanding challenges, including autonomous driving, robotics, remote sensing, and augmented reality. Despite this fact, the performance of 3D instance segmentation lags that of 2D instance segmentation, not least due to the additional challenges of 3D representation, and variable density of points.

\begin{figure}[htbp]
	\centering
	{
		\includegraphics[width=\columnwidth]{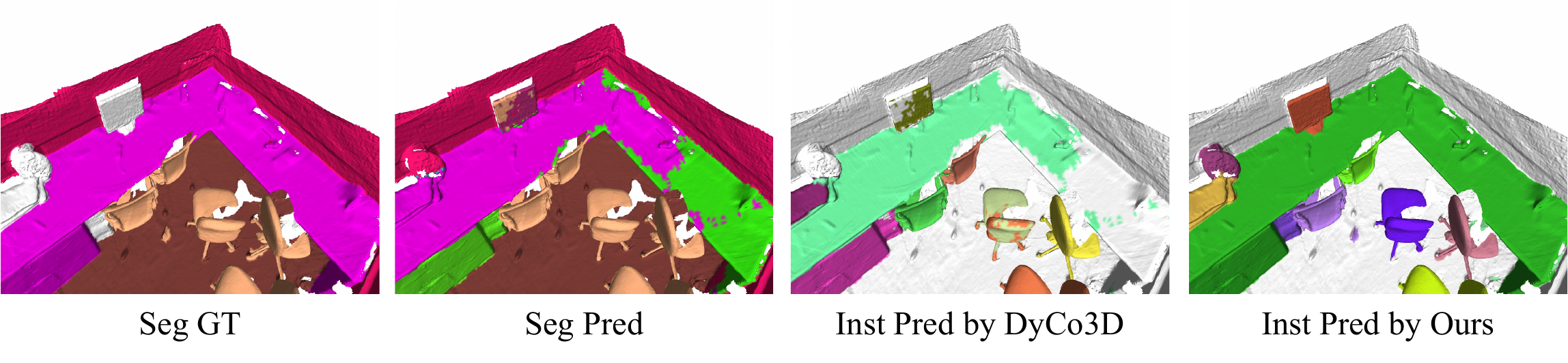}
	}

	\caption{
	A comparison of the instance segmentation results achieved by DyCo3D~\cite{He2021dyco3d} and our method. The subpar performance of instance segmentation for DyCo3D~\cite{He2021dyco3d} is caused by the dependency on semantic segmentation. Our method addresses the task in a per-point prediction fashion and removes the dependencies between different tasks of the model. Thus, it is free from the error accumulation introduced by the intermediate tasks. Best viewed in colors.
	}
	\label{fig:main}
\end{figure}

Most of the top-performing 3D instance segmentation approaches~\cite{jiang2020pointgroup,Engelmann20CVPR,He2021dyco3d,liang2021instance,chen2021hierarchical,hanoccuseg} involve a clustering step.
Despite their great success, clustering-based methods have their drawbacks: they are susceptible to the performance of the clustering approach itself, and its integration, due to either (1) error accumulation caused by the inter-task dependencies \cite{jiang2020pointgroup,He2021dyco3d,chen2021hierarchical} or (2) non-differentiable processing steps \cite{liang2021instance,hanoccuseg}.
For example, in PointGroup~\cite{jiang2020pointgroup}, instance proposals are generated by searching homogenous clusters that have identical semantic predictions and close centroid predictions. However, the introduced dependencies on both tasks make the results sensitive to the heuristics values chosen. DyCo3D~\cite{He2021dyco3d} addressed the issue by encoding instances as continuous functions. But the accuracy is still constrained by the semantic-conditioned convolution.
As a result, it can be impossible to recover from errors in intermediate stages, particularly given that many methods greedily associate points with objects (which leaves them particularly susceptible to early clustering errors).
Even with careful design, because of the diversity in the scales of instances, and the unbalanced distribution of semantic categories, the performance of these intermediate tasks is often far from satisfactory. This typically leads to fragmentation and merging of instances, as shown in Fig.~\ref{fig:main}.

In this paper, we remove the clustering step and the dependencies within the model and propose a much simpler pipeline working in a per-point prediction fashion. Every sampled point will generate a set of instance-related convolutional parameters, which are further applied for decoding the binary masks of the corresponding instances. However, building such a clustering-free and dependency-free pipeline is non-trivial. For example, removing the clustering step and conditional convolution in DyCo3D causes mAP to drop by more than 8\% and 6\%, respectively. 
We conduct comprehensive experiments and find the reason for the huge drop in performance is the ambiguity of the targets for the sampled points.
In 2D instance segmentation and object detection, the center prior, which assumes the predictions from the central areas of an instance are more likely to provide accurate results, offers a guideline to select well-behaved samples~\cite{tian2019fcos,tian2020conditional,ge2021ota}.
This distance-based prior is hard to apply in 3D, however, as the distribution of high-quality samples in 3D point clouds is irregular and unpredictable. The fact that objects can be arbitrarily close together in real 3D scenes adds additional complexity. Thus, the resulting ambiguity in point-instance associations can contaminate the training process and impact final performance.
Instead of applying a static or widely used distance-based strategy, we propose to optimally assign instances to samples via an Optimal Transport (OT) solution.
It is defined in terms of a set of suppliers and demanders, and the costs of transportation between them. We thus associated a demander with each instance prediction of the sampled point, and a supplier with each potential instance ground truth. The cost of transport reflects the affinity between each pair thereof.
The OT algorithm identifies the optimal strategy by which to supply the needs of each demander, given the cost of transport from each supplier.
The points will then be associated with the target corresponding to the demander to which it has allocated the greatest proportion of its supply.
The costs of transporting are determined by the Dice Coefficient, which is updated dynamically based on the per-point predictions. The OT solution not only minimizes the labor for heuristics tuning but allows it to make use of the sophisticated tools that have been developed for solving such problems. In particular, it can be efficiently solved by the off-the-shelf
Sinkhorn-Knopp Iteration algorithm~\cite{Marco2013_sinkhorn} with limited computation in training.

To summarise, our contributions are listed as follows.
\begin{itemize}
	\item We propose a clustering-free framework for 3D instance segmentation, working in a per-point prediction fashion. In doing so it removes the dependencies among different tasks and thus avoids error accumulation from the intermediate tasks.

	\item For the first time, we address the target assignment problem for 3D instance segmentation, which has been overlooked in the 3D community. Our proposed Optimal Transport solution is free from heuristics with improved accuracy.

	\item We achieve promising results on both ScanNet and S3DIS, with a much simpler pipeline. 
\end{itemize}

\section{Related Work}
\textbf{Target Assignment in 2D Images.}
The problem of associating candidates to targets arises commonly in 2D object detection.
Anchor-based detectors~\cite{renNIPS15fasterrcnn,lin2017retina,lin2017fpn} apply a hard threshold to an intersection-over-union measure to divide positive and negative samples. This approach can also be found in many other methods~\cite{cai18cascadercnn,he2018maskrcnn}.
Anchor-free detectors~\cite{tian2019fcos,zhou2019objects,kong2019foveabox} have drawn increasing attention due to their simplicity. These methods observe that samples around the center of objects are more likely to provide accurate predictions. Inspired by this center prior, some methods~\cite{tian2020conditional,tian2021fcos,kong2019foveabox,yu2016unitbox} introduce a classifier by treating these central regions as positive samples.
ATSS~\cite{zhang2020bridging}, in contrast, is adaptive in that it sets a dynamic threshold according to the statistics of the set of closest anchors. 
Free-Anchor~\cite{zhang2019freeanchor} frames
detector training as a maximum likelihood estimation (MLE) procedure and proposes a learning-based matching mechanism. Notably, OTA~\cite{ge2021ota} formulates the task of label assigning as Optimal Transport problem. 

\begin{figure*}[htbp]
	\begin{center}
		{
			\includegraphics[width=\columnwidth]{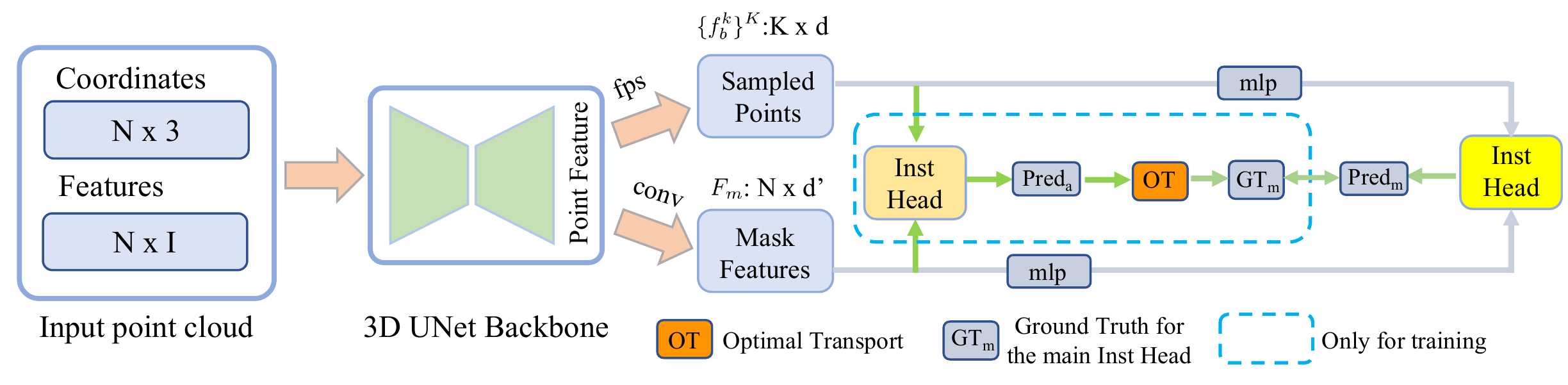}
			
		}
	\end{center}
	
	\caption{
		The framework of our proposed method. The `inst head' is designed to generate instance masks by applying dynamic convolution. K points are sampled via the farthest point sampling strategy.
		Each sampled point is responsible for one specific instance mask or background.
		The targets are calibrated dynamically via an Optimal Transport solution, which takes as input the mask prediction from the auxiliary head and outputs the calibrated ground truth for the main instance head. The targets for the auxiliary instance prediction `$\text{pred}_a$' are consistent with the instance label of the sampled points. 
	}
	\label{fig:framework}
\end{figure*}

\textbf{Instance Segmentation on 3D Point Cloud.} 
The task of instance segmentation in the 3D domain is complicated by the irregularity and sparsity of the point cloud. Unlike instance segmentation of images, in which top-down methods are the state-of-the-art, the leader board in instance segmentation of 3D point clouds has been dominated by bottom-up approaches due to unsatisfactory 3D detection results. 
SGPN~\cite{wang2018sgpn}, for instance, predicts an $N\times N$ matrix to measure the probability of each pair of points coming from the same instance, where $N$ is the number of total points. 
ASIS~\cite{wang2019asis} applies a discriminative loss function from~\cite{bra2017cvprdis} to learn point-wise embeddings. The mean-shift algorithm is used to cluster points into instances. Many works (\eg \cite{zhao2020jsnet,he2020eccvembedding,he2020eccvmemory,phamjsis3dcvpr19}) follow this metric-based pipeline. However, these methods often suffer from low accuracy and poor generalization ability due to their reliance on pre-defined hyper-parameters and complex post-processing steps. 
Interestingly, PointGroup~\cite{jiang2020pointgroup} exploits the voids between instances for segmentation. Both original and center-shifted coordinates are applied to search nearby points that have identical semantic categories. 
The authors of DyCo3D~\cite{He2021dyco3d} addressed the sensitivity of clustering methods to the grouping radius using dynamic convolution. Instead of treating clusters as individual instance proposals, DyCo3D utilized them to generate instance-related convolutional parameters for decoding masks of instances. 
Chen~\etal proposed HAIS~\cite{chen2021hierarchical}, which is also a clustering-based architecture. It addressed the problem of the over- and under-segmentation of PointGroup~\cite{jiang2020pointgroup} by deploying an intra-instance filtering sub-network and adapting the grouping radius according to the size of clusters. 
SSTN~\cite{liang2021instance} builds a semantic tree with superpoints~\cite{loicsuperpoint} being the leaves of the tree. The instance proposals can be obtained when a non-splitting decision is made at the intermediate tree node. A scoring module is introduced to refine the instance masks. 
\section{Methods}
The pipeline of the proposed method is illustrated in Fig.~\ref{fig:framework}, which is built upon a sparse convolution backbone~\cite{graham2018sparseconv}. It maintains a UNet-like structure and takes as input the coordinates and features, which have a shape of $N\times3$ and $N \times I$, respectively. $N$ is the total number of input points and $I$ is the dimension of input features. There is one output branch of mask features, which is used to decode binary masks of instances. It is denoted as $F_m \in \mathbb{R}^{N\times d'}$, where $d'$ is the dimension of the mask features. Inspired by DyCo3D~\cite{He2021dyco3d}, we propose to encode instance-related knowledge into a set of convolutional parameters and decode the corresponding masks with several 1$\times$1 convolutions. Different from DyCo3D, which requires a greedy clustering algorithm and a conditioned decoding step, our proposed method, on the other hand, removes the clustering step and the dependencies among different tasks, simplifying the network in a point-wise prediction pipeline.

\subsection{Preliminary on DyCo3D}
DyCo3D~\cite{He2021dyco3d} has three output branches: semantic segmentation, centroid offset prediction, and mask features. The breadth-first-searching algorithm is used to find out the homogenous points that have identical semantic labels and close centroid predictions. Each cluster is sent to the instance head and generates a set of convolution parameters for decoding the mask of the corresponding instance. Formally, the mask $\hat{M_k}$ predicted by the $k$-th cluster can be formulated as:
\begin{equation}
\begin{aligned}
\hat{M_k} &= Conv_{1x1}(feature, weight) \\
&=Conv_{1x1}(F_m\oplus C_\text{rel}^k, mlp(G(P_s, P_c)_k)) \odot \mathds{1}({P_s=s_k})
\end{aligned}
\label{eq:dyco3d_conv}
\end{equation}
The input features to convolution contains two parts: $F_m$ and $C_\text{rel}^k$. $F_m$ is the mask features shared by all instances.
$C_\text{rel}^k \in \mathbb{R}^{N\times 3}$ is the instance-specific relative coordinates, which are obtained by computing the difference between the center of the $k$-th cluster and all input points. $F_m$ and $C_\text{rel}^k$ are concatenated (`$\oplus$') along the feature dimension. 
The convolutional weights are dynamically generated by an mlp layer, whose input is the feature of the $k$-th cluster. The clustering algorithm $G(\cdot)$ takes the semantic prediction $P_s \in \mathbb{R}^N$ and centroid prediction $P_c \in \mathbb{R}^N$ as input and finds out a set of homogenous clusters. The $k$-th cluster is denoted as $G(\cdot)_k$. 
Besides, the dynamic convolution in DyCo3D is conditioned on the results of semantic segmentation. For example, DyCo3D can only discriminate one specific `Chair' instance from all points that are semantically categorized as `Chair', instead of the whole point set. 
It is implemented by an element-wise production (`$\odot$') with a binary mask (`$\mathds{1}(\cdot)$'). $s_k$ is the semantic label of the $k$-th cluster. 
Finally, the target mask for $\hat{M_k}$ is decided by the instance label of the $k$-th cluster.
More details can be found in \cite{He2021dyco3d}.

\subsection{Proposed Method}
Although promising, DyCo3D~\cite{He2021dyco3d} involves a grouping step to get the instance-related clusters, depending on the accuracy of semantic segmentation and offset prediction.
Besides, the conditional convolution also forces the instance decoding to rely on the results of semantic segmentation.
These inter-task dependencies cause error accumulation and lead to sub-par performance (See Fig.~\ref{fig:main}). 
In this paper, we propose a clustering-free and dependency-free framework in a per-point prediction fashion. Total $K$ points are selected via the farthest point sampling strategy. The instance head takes as input both the mask feature $F_m$ and point-wise feature $f_b^k$. 
The $k$-th mask $\hat{M_k}$ predicted by the instance head can be formulated as:
\begin{equation}
\begin{aligned}
\hat{M_k} &= Conv_{1x1}(feature, weight) \\
&=Conv_{1x1}(F_m\oplus C_\text{rel}^k, mlp(f_b^k))
\end{aligned}
\label{eq:ours_dyconv}
\end{equation}
where $f_b^k$ is the feature of the $k$-th sampled point from output of the backbone. $C_\text{rel}^k \in \mathbb{R}^{N\times3}$ is the relative position embedding, obtained by computing the difference between the coordinate of the $k$-th point and all other points. More details about the instance head can be found in supplementary materials.

However, building such a simplified pipeline is non-trivial. Removing the clustering step and conditional convolution causes the mAP of DyCo3D to drop dramatically. 
\begin{figure}[t]
	\centering
	{
		\includegraphics[height=4cm]{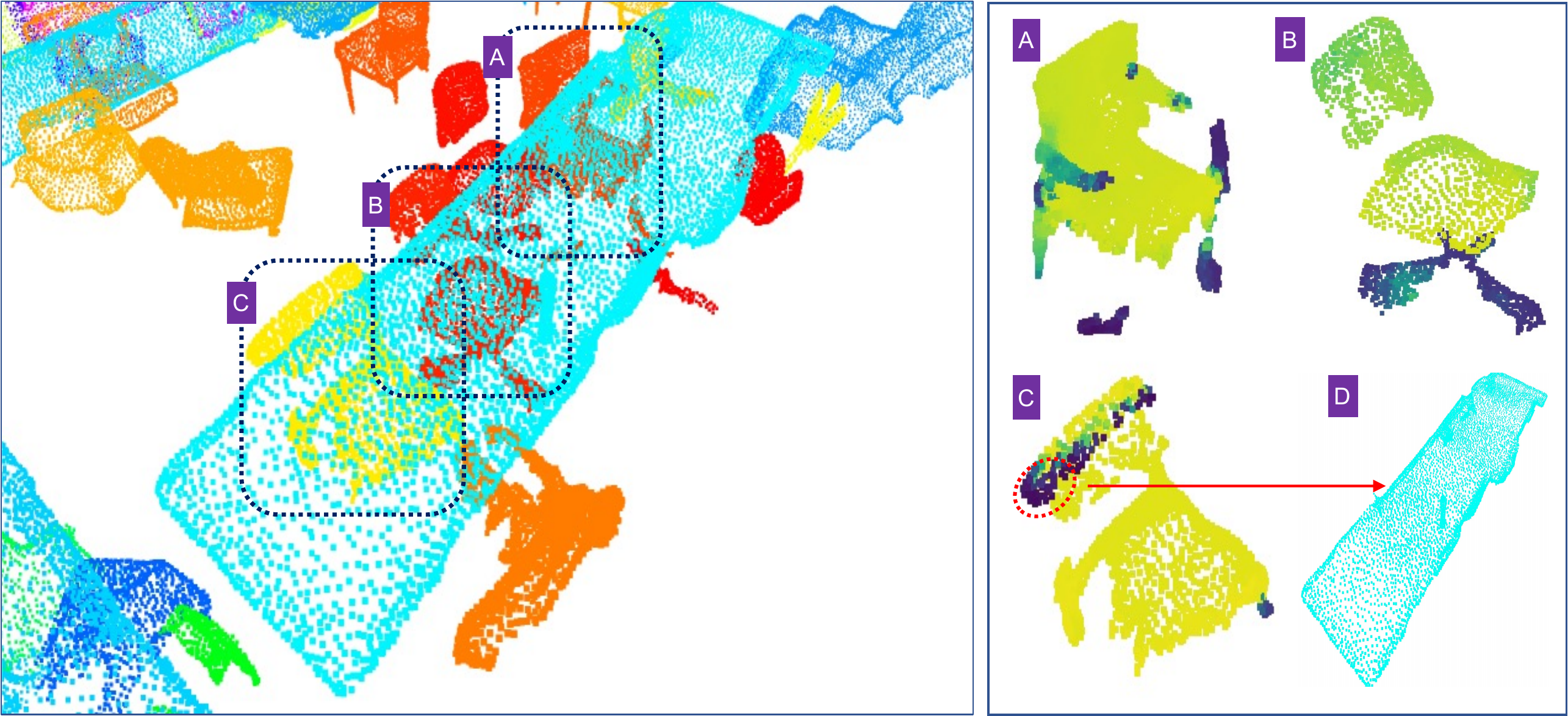}
	}
	
	\caption{ 
		The left image is an indoor scene with three instances of `Chair'. The right image is the quality of instance predictions by each point.
		The brighter the color, the more accurate the mask predicted by the point. Different from the 2D image, the distribution of the positive samples in 3D point cloud is irregular, making it hard to learn a criterion to select informative samples for each instance. In addition, the ambiguity of target assignment is widespread in the 3D scenes. Some samples in instance `C' show high-quality predictions of the instance `D'.
		Best viewed in color. 
	}
	\label{fig:ious}
\end{figure}

\subsubsection{Observation}
To find out the reasons that cause the failure of this point-wise prediction pipeline, we visualize the quality of masks predicted by each point (according to Eq.~\ref{eq:ours_dyconv}).
For training, the target mask for each point is consistent with its instance label.
As shown in Fig.~\ref{fig:ious}, the distribution of high-quality samples is irregular and can be influenced by many factors: (1) disconnection, (2) distance to the instance center, and (3) spatial relationships with other objects. 
Besides, the fact that objects can be arbitrarily close together in real 3D scenes adds additional complexity. As illustrated in Fig.~\ref{fig:ious}(c,d), the poorly behaved samples in `chair c' can accurately predict the mask of the `desk'. Such ambiguity introduced by the static assigning strategy contaminates the training process, leading to inferior performance.  

\subsubsection{Target Assignment}
Although the task of target assignment has shown its significance in 2D object detection and instance segmentation~\cite{zhang2019freeanchor,zhang2020bridging,ge2021ota}, to the best of our knowledge, there is very little research in the 3D domain. One of the most straightforward ways is to define a criterion to select a set of informative samples for each instance.
For example, thanks to the center prior~\cite{tian2019fcos}, many approaches~\cite{tian2020conditional,zhou2019objects,kong2019foveabox,yu2016unitbox} in the 2D domain treat the central areas of the instance as positive candidates. However, such regularity is hard to define for the 3D point cloud, as shown in Fig.~\ref{fig:ious}. Quantitative results can be found in Tab.~\ref{tab:ablation_study}. 

Instead of applying a static strategy or learning an indicative metric, we propose to assign a suitable target for each sample based on its prediction. A background mask (\ie all zeros) is added to the target set to address the poorly-behaved points. 

\subsubsection{Optimal Transport Solution}
Given $K$ sampled points (via farthest point sampling) and their corresponding mask predictions $\{\hat{M_k} \}^K$ (using Eq.~\ref{eq:ours_dyconv}), the goal of target assignment is to find a suitable target for each prediction in training. There are T+1 targets in total, including T instance masks $\{M_t\}^T$ and one background mask $M_\text{T+1} $ (zero mask).
Inspired by \cite{ge2021ota}, we formulate the task as an Optimal Transport problem, which seeks a plan by transporting the `goods' from
suppliers (\ie Ground Truth and Background Mask) to demanders (\ie predictions of the sampled points) at a minimal transportation cost.

Supposing the $t$-th target has $\mu_t$ unit of goods and each prediction needs one unit of goods, we denote the cost for transporting one unit of goods from the $t$-th target to the $k$-th prediction as $ C_{tk}$. By applying Optimal Transport, the task of the target assignment can be written as:

\begin{equation}
\begin{aligned}
&{ \bm U}^* =  \mathop{\arg\min}_{ \bm U \in \mathbb{R} ^{(T+1) \times K}_{+}} \sum_{t,k} { C}_{tk} { U}_{tk} \\
&\text{s.t.} \quad {\bm U} {\bm 1}_{K} = \bm \mu_{T+1}, \ {\bm U}^\mathsf{T} {\bm 1}_{T+1} = {\bm 1}_{K},
\end{aligned}
\label{eq:ot_problem}
\end{equation}
where $\bm{U}^* $ is the optimal assignment plan, ${U}_{tk}$ is the amount of labels transported from the $t$-th target to the $k$-th prediction. $\bm \mu_{T+1}$ is the label vector for all $T+1$ targets. 
The transportation cost $ C_{tk}$ is defined as:
\begin{equation}
C_{tk} = 
\begin{cases}
\mathcal{L}_\text{dice}(M_t, \hat{M_k}) 
& t\leq T \\
\mathcal{L}_\text{dice}(1-M_t, 1- \hat{M_k})
& t = T+1
\end{cases}
\label{eq:ot}
\end{equation}
where $\mathcal{L}_\text{dice}$ denotes the dice loss. To calculate the cost between the background target and the prediction, we use $1-M_t$ and $1-\hat{M_k}$ for a numerically stable training. The restriction in Eq.~\ref{eq:ot_problem} describes that (1) the total supply must be equal to the total demand and (2) the goods demand for each prediction is 1 (\ie each prediction needs one target mask). 
Besides, the label vector $\bm \mu_{T+1}$, indicating the total amount of goods held by each target, is updated by: 

\begin{equation}
\mu_t = 
\begin{cases}
int(\sum_k IoU(\hat{M_k}, M_t) )& t\leq T \\
K-\sum_{i=1}^T \mu_i& t = T+1
\end{cases}
\label{eq:mu_t}
\end{equation}
where $\mu_{T+1}$ refers to the target amount maintained in the background target and $int(\cdot$) is the rounding operation. 
According to Eq.~\ref{eq:mu_t}, the amount of supplied goods for each target is dynamically decided, depending on its IoU with each prediction. Due to the restriction in Eq.~\ref{eq:ot_problem}, we set $\mu_{T+1}$ equal to $K-\sum_{t=1}^T $.
The efficient Sinkhorn-Knopp algorithm~\cite{Marco2013_sinkhorn} allows it to obtain $\bm U^*$ with limited computation overhead.
After getting the optimal assignment $\bm U^*$, the calibrated targets for the $K$ sampled points can be determined by assigning each point with the target that transports the largest amount of goods to it. The details of the algorithm are in the supplementary materials. 

Compared with \cite{ge2021ota}, the number of the demanders is much fewer. Thus, the minimum supply of each target can be zero in training. Doing so may make the model fall into a trivial solution when $K$ is small: all predictions are zero masks and assigned to the background target due to the lowest transportation cost in Eq.~\ref{eq:ot}. To this end, we propose a simple yet effective way by introducing an auxiliary instance head, whose targets are consistent with the instance labels of the sampled points. We use the predictions from this auxiliary head to calculate the cost matrix in Eq.~\ref{eq:ot}. The dynamically calibrated targets are used for the main instance head. 
To alleviate the impact of the wrongly assigned samples in the auxiliary head, the loss weight for this auxiliary task is decreasing in training.

\subsection{Training}

To summarize, the loss function includes two terms for training, including the auxiliary loss term $\mathcal{L}_\text{a}$ and the main task loss term $\mathcal{L}_\text{m}$:
\begin{equation}
\mathcal{L} = w_a\sum_{k=1}^{K}\mathcal{L}_\text{a}(M^a_\text{k}, \hat{M}^a_\text{k}) + \sum_{k=1}^{K}\mathcal{L}_\text{m}(M^m_\text{k}, \hat{M}^m_\text{k}) 
\label{eq:loss}
\end{equation}
where $\{M^a_\text{k}\}^K \in \{0,1\}^{K\times N}$ is the ground truth masks for the $K$ predictions. These targets are static and decided by the instance labels of the $K$ sampled points. $\{M^m_\text{k}\}^K \in \{0,1\}^{K\times N}$ is the set of the calibrated targets for the main instance head. $\{\hat{M}^a_\text{k}\}^K$ and $\{\hat{M}^m_\text{k}\}^K$ are the predictions from auxiliary and main instance heads, respectively. $w_a$ is the loss weight for the auxiliary task. We set $w_a$ to 1.0 with a decaying rate of 0.99. Early in the training phase, the static targets for the auxiliary task play a significant role in stabilizing the learning process. The loss of the main task is involved until the end of a warming-up period, which is set to 6k steps. 
So far, we have obtained a set of binary masks. There are many ways to obtain the corresponding categories, for example, adding a classification head for each mask proposal. In our paper, we implement it by simply introducing a semantic branch. The category $c_k$ of the $k$-th instance is the majority of the semantic predictions within the foreground mask of $\hat{M}^m_k$. Instances with a number of points less than 50 are ignored. 

\section{Experiments}
We conduct comprehensive experiments on two standard benchmarks to validate the effectiveness of our proposed method: ScanNet~\cite{dai2017scannet} and Stanford 3D Indoor Semantic Dataset (S3DIS)~\cite{armeni2016s3dis}. 
\subsection{Datasets}
ScanNet has 1613 scans in total, which are divided into training, validation, and testing with a size of 1201, 312, and 100, respectively. The task of instance segmentation is evaluated on 18 classes. Following ~\cite{He2021dyco3d}, we report the results on the validation set for ablation study and submit the results on the testing set to the official evaluation server. The evaluation metrics are mAP (mean average precision ) and AP@50.

S3DIS contains more than 270 scans, which are collected on 6 large indoor areas. It has 13 categories for instance segmentation. Following the previous method~\cite{wang2019asis}, the evaluation metrics include: mean coverage (mCov), mean weighed coverage (mWCov), mean precision (mPrec), and mean recall (mRec).

\begin{table}[!t]
	\normalsize 
	\centering

	\begin{center}
		\begin{tabular}{c|ccc|c|c|c}
			\toprule
			
			Method & CP &DT &AUX  & mAP &AP@50 &AP@25  \\
			\toprule

			\text{Baseline} & & & &33.7 &52.4 &65.0 \\
			&\checkmark& &  & 34.1 &53.2 & 65.4 \\ 
			\midrule
			& &\checkmark & & 36.8 & 54.8 & 65.9 \\
			& & &\checkmark & 36.5 & 54.3 & 65.7 \\
			\textbf{Ours}   & &\checkmark &\checkmark &39.6 &59.2 &70.4 \\ 
			
			\bottomrule
		\end{tabular}
	\end{center}
	\caption{Component-wise analysis on ScanNetV2 validation set. \textbf{CP}: the center prior tailored for 3D point cloud. \textbf{DT}: dynamic targets assignment using Optimal Transportation. \textbf{AUX}: the auxiliary loss used in Eq.~\ref{eq:loss}. %
	}
	
	\label{tab:ablation_study}
\end{table}
\subsection{Implementation Details}
\label{sec:exp_setting}
The backbone model we use is from \cite{graham2018sparseconv}, which maintains a symmetrical UNet structure. It has 7 blocks in total and the scalability of the model is controlled by the channels of the block. To prove the generalization capability of our proposed method, we report the performance with both small and large backbones, denoted as $\textbf{\text{Ours-S}}$ and $\textbf{\text{Ours-L}}$, respectively. The small model has a channel unit of 16, while the large model is 32. The default dimension of the mask features is 16 and 32, respectively.

For each input scan, we concatenate the coordinates and RGB values as the input features. All experiments are trained for 60K iteration with 4 GPUS. The batch size for each GPU is 3. The learning rate is set to 1e-3 and follows a polynomial decay policy. In testing, the computation related to the auxiliary head is ignored. 
Only Non-Maximum-Suppression (NMS) is required to remove the redundant mask predictions for inference, with a threshold of 0.3.

\subsection{Ablation Studies}
In this section, we verify the effectiveness of the key components in our proposed method. For a fair comparison, all experiments are conducted on the validation set of ScanNet~\cite{dai2017scannet} with the smaller model. 

\textbf{Baseline.} 
We build a strong baseline by tailoring CondInst~\cite{tian2020conditional} for the 3D point cloud. It works in a per-point prediction fashion and each sampled point has a static target, which is consistent with the corresponding instance label.
As shown in Tab.~\ref{tab:ablation_study}, our method achieves 33.7\% 52.4\%, and 65.0\% in terms of mAP, AP@50, and AP@25, respectively.
With a larger number of sampled points and longer iterations, our baseline model surpasses the implementation of DyCo3D~\cite{He2021dyco3d} by a large margin. 

\textbf{Center Prior in 3D.} 
To demonstrate the difficulty of selecting informative samples in 3D, we tailor the center prior~\cite{tian2019fcos} to 3D point cloud. As points are collected from the surface of the objects, centers of 3D instances are likely to be in empty space. To this end, we first predict the offset between each point and the center of the corresponding object. If the distance between the center-shifted point and the ground truth is close ($\le 0.3$m), the point is regarded as positive and responsible for the instance. If the distance is larger than 0.6m, the point is defined as negative. Other points are ignored for training. As presented in Tab.~\ref{tab:ablation_study}, selecting positive samples based on the 3D center prior only boosts 0.4\% and 0.8\% in terms of mAP and mAP@50, respectively. The incremental improvement demonstrates the difficulty of selecting informative samples in 3D. In contrast, we propose to apply a dynamic strategy, by which the target for each candidate is determined based on its prediction.

\begin{figure}
	\centering
	{
		\includegraphics[width=7cm]{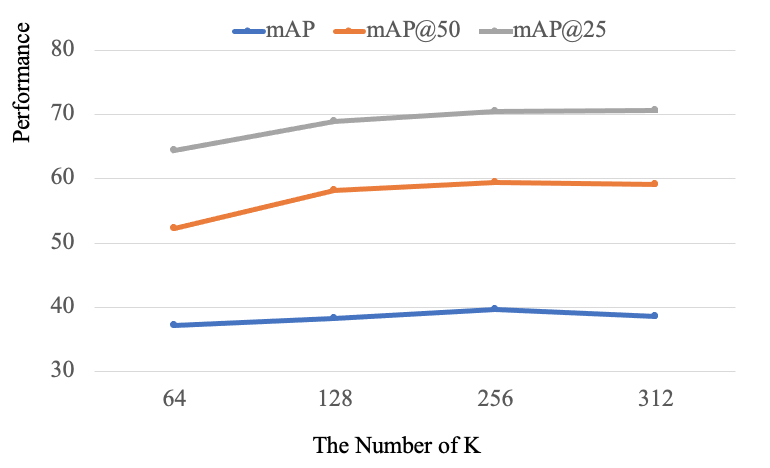}
	}

	\caption{Ablation study on the number of the sampling point. 
	}
	
	\label{fig:k_variation}
\end{figure}

\textbf{Dynamic Targets.} To show the effectiveness of the dynamic strategy, we implement an experiment by removing the auxiliary head. As the predictions are basically random guesses in the early stage of the training, we first warm up the model for 12k iterations with a static assignment to avoid the trivial solution. In the remaining steps, targets are calibrated by the Optimal Solution. As shown in Tab.~\ref{tab:ablation_study}, our approach boosts the performance of the baseline model by 3.1\%, 2.4\%, and 0.9\%, in terms of mAP, AP@50, and AP@25, respectively.

\begin{table}[!t]
    \normalsize 
	\centering

	\begin{center}
		\begin{tabular}{c c}
			\toprule
			
			\multicolumn{2}{c}{\textbf{3D Object Detection}}  \\ 
			\midrule
			ScanNetV2      & AP@50\%  \\
			\midrule
			
			MRCNN 2D-3D~\cite{he2018maskrcnn} & 10.5 \\
			F-PointNet~\cite{qi2017frustum} & 10.8 \\
			GSPN~\cite{yi2018gspn}        & 17.7 \\
			3D-SIS~\cite{hou20193dsis}    & 22.5 \\
			VoteNet~\cite{qi2019deep}      & 33.5 \\
			PointGroup~\cite{jiang2020pointgroup} &42.3 \\
			DyCo3D~\cite{He2021dyco3d}     &{45.3} \\ 
			3D-MPA~\cite{Engelmann20CVPR} & 49.2\\
			Ours &\textbf{51.0}\\
			
			\bottomrule
		\end{tabular}
	\end{center}
	
	\caption{The performance of 3D object detection, tested on ScanNet validation set. AP@50 is reported.
	}
	
	\label{tab:object_detection}
\end{table}

\begin{figure*}[!h]
	\centering
	{
		\includegraphics[width=\columnwidth]{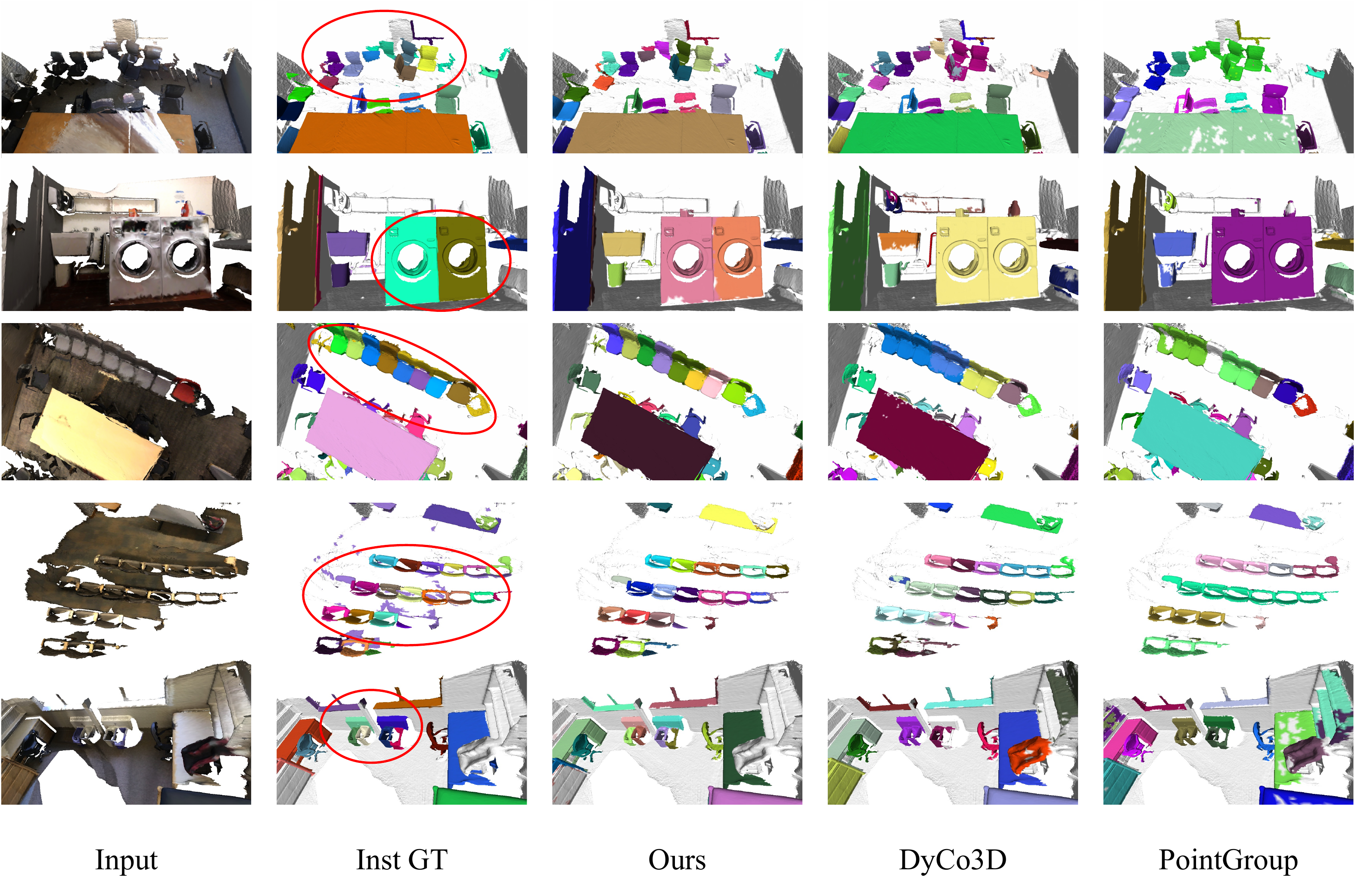}
	}
	
	\caption{Comparison with the results of DyCo3D~\cite{He2021dyco3d} and PointGroup \cite{jiang2020pointgroup}. The ellipses highlight specific over-segmentation/joint regions. Instances are presented with different colors. Best viewed in color.}
	\label{fig:results}
\end{figure*}
\textbf{Auxiliary Supervision.}
As illustrated in Fig.~\ref{fig:framework}, we propose to regularize the intermediate layers by introducing an auxiliary instance head for decoding the instance masks. The targets for this task are static and consistent with the instance labels. Besides, as the generated parameters are convolving with the whole point set, large context and instance-related knowledge are encoded in the point-wise features. To remove the influence of the dynamic assignment, both auxiliary and the main task are applying a static assignment strategy. As shown in Tab.~\ref{tab:ablation_study},
the auxiliary supervision brings 2.8\%, 1.9\%, and 0.7\% improvement in terms of mAP, mAP@50, and mAP@25, respectively. In addition to the encoded large context, the predicted instance masks are also applied to the Optimal Solution to obtain calibrated targets. Combining with the proposed dynamic assignment strategy, it further boosts mAP, AP@50, and AP@25 for 3.1\%, 4.4\%, and 4.5\%, respectively, achieving 39.6\% in terms of mAP with a small backbone.

\textbf{Analysis on Efficiency.} Our method takes the whole scan as input, without complex pre-processing steps. Similar to DyCo3D~\cite{He2021dyco3d}, the instance head is implemented in parallel. To make a fair comparison, we set K equal to the average number of clusters in DyCo3D. Using the same GPU, the mAP of our proposed method is 1.8\%  higher than DyCo3D and the inference time is 26\% faster than DyCo3D. 

\begin{table}[!t]
	\normalsize
	\begin{center}
		\begin{tabular}{c|c|c|c|c}
			\toprule
			Method  & mCov    & mWCov    &  mPrec & mRec \\
			
			\midrule
			\multicolumn{5}{c}{Test on Area 5} \\
			\midrule
			SGPN'18 ~\cite{wang2018sgpn}   &  32.7  & 35.5  &  36.0  & 28.7   \\
			
			ASIS'19 ~\cite{wang2019asis}   & 44.6  &  47.8 &  55.3 &  42.4 \\
			3D-BoNet'19 ~\cite{yang20193dbonet}   & -  &  - &  57.5 &  40.2 \\
			3D-MPA'20~\cite{Engelmann20CVPR}   & - &- &63.1 &58.0\\
			MPNet'20~\cite{he2020eccvmemory}      & {50.1} &{53.2} &{62.5} &{49.0} \\
			
			InsEmb'20~\cite{he2020eccvembedding}   &49.9 & 53.2 & 61.3 &48.5 \\
			PointGroup'20~\cite{jiang2020pointgroup}  &- &- & 61.9 & 62.1 \\
			DyCo3D'21~\cite{He2021dyco3d}  &{63.5} &{64.6} &{64.3} &{64.2} \\
			
			HAIS'21~\cite{chen2021hierarchical} & \textbf{64.3} &\textbf{66.0} &71.1 &65.0\\
			SSTNet'21~\cite{liang2021instance} &- &- &65.5 &64.2 \\
			\textbf{Ours} &\textbf{64.3} & 65.3 & \textbf{73.1} & \textbf{65.2}\\
			\midrule
			\multicolumn{5}{c}{Test on 6-fold} \\
			\midrule
			SGPN'18 ~\cite{wang2018sgpn} &  37.9  & 40.8  &  31.2  & 38.2   \\
			MT-PNet'19 ~\cite{phamjsis3dcvpr19}  &-   &-   &24.9 &- \\
			MV-CRF'19 ~\cite{phamjsis3dcvpr19} &- &- &36.3 &- \\
			ASIS'19 ~\cite{wang2019asis}&51.2 &55.1  & 63.6 & 47.5 \\
			3D-BoNet'19 ~\cite{yang20193dbonet} &- &- &65.6 &47.6 \\
			PartNet'19 ~\cite{Mo_2019_CVPR}  &- &- &56.4 &43.4 \\ 
			InsEmb'20\cite{he2020eccvembedding} &54.5 & 58.0 &67.2 &51.8\\
			MPNet'20 \cite{he2020eccvmemory}  &{55.8} &{59.7} &{68.4} &{53.7} \\
			PointGroup'20 \cite{jiang2020pointgroup} &- &- &69.6 &69.2 \\
			3D-MPA'20 \cite{Engelmann20CVPR} &- &- &66.7 & 64.1 \\
			HAIS'21~\cite{chen2021hierarchical} & 67.0 &70.4 &73.2 &69.4\\
			SSTNet'21~\cite{liang2021instance} &- &- &73.5 &73.4 \\
			
			\textbf{Ours} &\textbf{71.5}  &\textbf{74.1}  &\textbf{76.4}  &\textbf{74.0} \\
			
			\bottomrule
		\end{tabular}
	\end{center}
	
	\caption{Instance segmentation results on S3DIS. The performance on both Area-5 and 6-fold cross-validation is reported.
	}
	
	\label{tab:s3dis_ins_results}
\end{table}

\textbf{Number of Random Selected Samples.}
We randomly select $K$ points, each of which is responsible for one specific instance or the background (all zeros). In this part, we study the influence of the value of $K$. The performance is shown in Fig.~\ref{fig:k_variation}. We set K to 256 for its highest mAP.

\textbf{The Dimension of the Mask Feature.}
The mask feature contains the knowledge of instances. We conduct experiments to show the influence of different dimensions of the mask feature. 
We find the fluctuation of the performance is relatively small when the dimension is greater than 8, showing the strong robustness of our method to the variation of $d'$. We set $d'$ to 16 in our experiments.

\begin{table*}[!h]
	\resizebox{\textwidth}{!}{
		\begin{tabular}{l|cc|cccccccccccccccccc}
			\toprule
			&\textbf{AP@50} &\textbf{mAP}& \rotatebox{90}{cabinet} & \rotatebox{90}{bed} & \rotatebox{90}{chair} & \rotatebox{90}{sofa} & \rotatebox{90}{table} & \rotatebox{90}{door} & \rotatebox{90}{window} & \rotatebox{90}{bookshe.} & \rotatebox{90}{picture} & \rotatebox{90}{counter} & \rotatebox{90}{desk} & \rotatebox{90}{curtain} & \rotatebox{90}{fridge} & \rotatebox{90}{s.curtain} & \rotatebox{90}{toilet} & \rotatebox{90}{sink} & \rotatebox{90}{bath} & \rotatebox{90}{otherfu.} \\
			\midrule

			SGPN~\cite{wang2018sgpn}&11.3 &- &10.1&16.4&20.2&20.7&14.7&11.1&11.1&0.0&0.0&10.0&10.3&12.8&0.0&0.0&48.7&16.5&0.0&0.0\\
			3D-SIS~\cite{hou20193dsis}&18.7&- &19.7&37.7&40.5&31.9&15.9&18.1&0.0&11.0&0.0&0.0&10.5&11.1&18.5&24.0&45.8&15.8&23.5&12.9\\
			3D-MPA~\cite{Engelmann20CVPR}&{59.1} &35.3 &{51.9}&{72.2}&{83.8}&{66.8}&{63.0}&{43.0}&{44.5}&{58.4}&{38.8}&{31.1}&{43.2}&
			{47.7}&{61.4}&\textbf{80.6}&{99.2}&{50.6}&{87.1}&{40.3}\\
			PointGroup~\cite{jiang2020pointgroup} &56.9 &34.8 &48.1 &69.6 &87.7 &71.5 &62.9 &42.0 &46.2 &54.9 & 37.7 &22.4 &41.6 & 44.9&37.2 &64.4 &98.3 &61.1 &80.5 &53.0 \\
			
			DyCo3D-S~\cite{He2021dyco3d}&57.6 &35.4 &50.6 &{73.8} &84.4 & 72.1 &{69.9} &40.8 &44.5 &\textbf{62.4} & 34.8 &21.2 & 42.2 &37.0 & 41.6 & 62.7 &92.9 &61.6 & 82.6 &47.5 \\

			HAIS-S~\cite{chen2021hierarchical}  &59.1 &38.0 &54.4 &76.0 &87.7 &69.4 & 66.5 & 47.5 & 48.5 & 53.1 & 43.6 &24.0 & 50.9 & \textbf{55.8} & 45.1 & 58.5 &94.7 & 53.6 & 80.8 & 53.0 \\
			
			\textbf{Ours-S} & 59.2&39.6  &51.1 & 75.9 & 86.5 & \textbf{72.8} & 67.3 & 45.2 & \textbf{52.3} & 57.2 & 43.8 & 25.7 & 40.5 & 53.7 & 37.2 & 59.4 & 98.2 & 58.9 & 87.0 &52.9 \\
			\midrule
			
			DyCo3D-L~\cite{He2021dyco3d} & {61.0} & {40.6} &{52.3} & 70.4 &{90.2} & 65.8 &69.6 & 40.5 & {47.2} &48.4 &{44.7} & {34.9} & {52.3} &47.5 &51.5 & 70.3 & 94.8 & \textbf{74.3} & 77.4 &{56.4} \\
			
			HAIS-L~\cite{chen2021hierarchical} & \textbf{64.0} & 43.5  &55.4 & 70.2 &82.5 &67.7 & 75.3 & 48.1 & 51.5 & 49.4 & \textbf{48.7} & \textbf{47.8} & \textbf{58.5} & 55.7 & \textbf{53.0} &76.1 & \textbf{100.0} & 69.2 & \textbf{87.1} & 56.3 \\ 
			
			\textbf{Ous-L}  &63.7 & \textbf{45.6} &\textbf{58.5} &\textbf{78.5}  &\textbf{93.6} &63.2 &\textbf{76.5}  &\textbf{55.6}  &48.5  &59.4  &38.3  &36.9  &54.2  &50.7  &46.2  &72.3 &98.3  &68.8  &\textbf{87.1}  &\textbf{59.5}  \\ 

			\bottomrule
		\end{tabular}
	}
	\caption{Quantitative comparison on the validation set of ScanNetV2. To make a fair comparison, we report the performance with different model scalability. The performance of HAIS-S is obtained by using the official training code. 
	}
	\label{tab:scannet_val}
\end{table*}

\begin{table*}[!t]
	\centering
	\resizebox{\textwidth}{!}{
		\begin{tabular}{l|c|cccccccccccccccccc}
			\toprule
			&\textbf{mAP}& \rotatebox{90}{bathtub} & \rotatebox{90}{bed} & \rotatebox{90}{bookshe.} & \rotatebox{90}{cabinet} & \rotatebox{90}{chair} & \rotatebox{90}{counter} & \rotatebox{90}{curtain} & \rotatebox{90}{desk} & \rotatebox{90}{door} & \rotatebox{90}{otherfu.} & \rotatebox{90}{picture} & \rotatebox{90}{refrige.} & \rotatebox{90}{s.curtain} & \rotatebox{90}{sink} & \rotatebox{90}{sofa} & \rotatebox{90}{table} & \rotatebox{90}{toilet} & \rotatebox{90}{window} \\
			\midrule

			R-PointNet~\cite{yi2018gspn} &15.8 & 35.6 & 17.3 & 11.3 & 14.0 & 35.9 & 1.2 & 2.3 & 3.9 & 13.4 & 12.3 & 0.8 & 8.9 & 14.9 & 11.7 & 22.1 & 12.8 & 56.3 & 9.4 \\
			3D-SIS~\cite{hou20193dsis} &16.1 & 40.7 & 15.5 & 6.8 & 4.3 & 34.6 & 0.1 & 13.4 & 0.5 & 8.8 & 10.6 & 3.7 & 13.5 & 32.1 & 2.8 & 33.9 & 11.6 & 46.6 & 9.3 \\
			MASC~\cite{Liu2019masc} &25.4 & 46.3 & 24.9 & 11.3 & 16.7 & 41.2 & 0.0 & 37.4 &7.3 & 17.3 & 24.3 & 13.0 & 22.8 & 36.8 &16.0 & 35.6 & 20.8 & 71.1 & 13.6\\
			PanopticFusion~\cite{Narita2019iros}  &21.4 & 25.0 & 33.0 & 27.5 & 10.3 & 22.8 & 0.0 & 34.5 & 2.4 & 8.8 & 20.3 & 18.6 & 16.7 & 36.7 & 12.5 & 22.1 & 11.2 & 66.6 & 16.2  \\
			3D-BoNet~\cite{yang20193dbonet} &25.3 & 51.9 & 32.4 & 25.1 & 13.7 & 34.5 & 3.1 & 41.9 & 6.9 & 16.2 & 13.1 & 5.2 & 20.2 & 33.8 & 14.7 & 30.1 & 30.3 & 65.1 &17.8 \\
			MTML~\cite{Jean2019mtml} &28.2 & 57.7 & 38.0 & 18.2 & 10.7 & 43.0 & 0.1 & 42.2 & 5.7 & 17.9 & 16.2 & 7.0 & 22.9 & 51.1 & 16.1 & 49.1 & 31.3 & 65.0 & 16.2\\
			3D-MPA~\cite{Engelmann20CVPR} &35.5 &45.7 & 48.4 & 29.9 & 27.7 & 59.1 & 4.7 & 33.2 & 21.2 & 21.7 & 27.8 & 19.3 & 41.3 & 41.0 &19.5 &57.4 & 35.2 & 84.9 & 21.3  \\
			DyCo3D~\cite{He2021dyco3d}&39.5 &64.2 & 51.8 & 44.7 & 25.9 & 66.6 & 5.0 & 25.1 & 16.6 & 23.1 & 36.2 & 323.2 & 33.1 & 53.5 & 22.9 & {58.7} & 43.8 & 85.0 & 31.7 \\
			PointGroup~\cite{jiang2020pointgroup} &40.7 & 63.9 & 49.6 & 41.5 & 24.3 & 64.5 & 2.1 & \textbf{57.0} & 11.4 & 21.1 & 35.9 & 21.7 &42.8 &66.0 & 25.6 & 56.2 & 34.1 & 86.0 &29.1  \\
			HAIS~\cite{chen2021hierarchical} &{45.7} &70.4 &\textbf{56.1} &{45.7} & \textbf{36.4} &67.3 & 4.6 & 54.7 & 19.4 & 30.8 & {42.6} &{28.8} & {45.4} & {71.1} & {26.2} & 56.3 & 43.4 & 88.9 &\textbf{34.4}\\
			\textbf{Ours} &43.8 &{81.5} &50.7 &33.8 &35.5 &\textbf{70.3} &{8.9} &39.0 & \textbf{20.8}&{31.3} &37.3 &{28.8} &40.1 &66.6 &24.2 &55.3 &{44.2} &{91.3} &29.3\\
			\midrule
			OccuSeg$^*$~\cite{hanoccuseg} & 44.3 &\textbf{85.2} & 56.0 & 38.0 & 24.9 & 67.9 & 9.7 & 34.5 & 18.6 & 29.8 & 33.9 & 23.1 & 41.3 & 80.7 & 34.5 & 50.6 & 42.4 &\textbf{97.2} & 29.1 \\
			SSTN$^*$~\cite{liang2021instance} & \textbf{50.6} & 73.8 & 54.9 & \textbf{49.7} & 31.6 & 69.3 & \textbf{17.8} & 37.7 &19.8 & \textbf{33.0} & \textbf{46.3} & \textbf{57.6} & \textbf{51.5} & \textbf{85.7} &\textbf{49.4} &\textbf{63.7} & 45.7 & 94.3 & 29.0\\
			
			\bottomrule[1pt]
			
		\end{tabular}
	}
	
	\caption{Quantitative results on ScanNetV2 testing set. \textbf{The last two methods are relying on complex preprocessing algorithms to obtain superpoints, which is time-consuming.}
	}
	
	\label{tab:scannet_test}
\end{table*}

\subsection{Comparison with State-of-the-art Methods}
We compare our method with other state-of-the-art methods on both S3DIS and ScanNet datasets.

\textbf{3D Detection.}
Following~\cite{He2021dyco3d,Engelmann20CVPR}, we evaluate the performance of 3D detection on the ScanNet dataset. The results are obtained by fitting axis-aligned bounding boxes for predicted masks, as presented in Tab.~\ref{tab:object_detection}. Our method surpasses DyCo3D~\cite{He2021dyco3d} and 3D-MPA~\cite{Engelmann20CVPR} by 4.8\% and 1.8\% in terms of mAP, respectively. The promising performance demonstrates the compactness of the segmentation results.

\textbf{Instance Segmentation on S3DIS.}
Following the evaluation protocols that are widely applied in the previous approaches, experiments are carried out on both Area-5 and 6-Fold cross-validation. 
As shown in Tab.~\ref{tab:s3dis_ins_results}, our proposed method achieves the highest performance and surpasses previous methods with a much simpler pipeline. With 6-fold validation, our method improves HAIS~\cite{chen2021hierarchical} by 4.5\%, 3.7\%, 3.2\%, and 4.6\% in terms of mConv, mWConv, mPrec, and mRec, respectively.
The proposed approach works in a fully end-to-end fashion, removing the error accumulation caused by the inter-task dependencies.

\textbf{Instance Segmentation on ScanNet.}
The performance of instance segmentation on the validation and testing sets of ScanNet~\cite{dai2017scannet} is reported in Tab.~\ref{tab:scannet_val}
and Tab.~\ref{tab:scannet_test}, respectively. 
On the validation set, we report the performance with both small and large backbones, denoted as $\text{Ours-S}$ and $\text{Ours-L}$, respectively. It surpasses previous top-performing methods on both architectures in terms of mAP, demonstrating strong generalization capability. Compared with DyCo3D~\cite{He2021dyco3d}, our approach exceeds it by 4.2\% in terms of mAP. The qualitative result is illustrated in Fig.~\ref{fig:results}. We also make a fair comparison with HAIS~\cite{chen2021hierarchical}, the highest mAP is achieved on the validation set.

\section{Conclusion and Future Works}
In this paper, we propose a novel pipeline for 3D instance segmentation, which works in a per-point prediction fashion and thus removes the inter-task dependencies. We show that the key to its success is the target assignment, which is addressed by an Optimal Transport solution. Without bells and whistles, our method achieves promising results on two commonly used datasets.

The sampling strategy used in our method is fps, which is slightly better than random sampling. We believe there exist other informative strategies that can further improve the performance.
In addition, due to the continuity representation capability, our method offers a simple solution to achieve instance-level reconstruction with the sparse point cloud. We leave these for future works.

\clearpage

\appendix
\section{Details of the Instance HEAD}
Given both instance-related filters and the position embedded features, we are ready to decode the masks of instances. The filters for the $k$-th instance are generated by the point feature $f_b^k$. The position embedded features have a dimension of $d'+3$, including the mask feature $F_m$ and the relative coordinate feature $C_\text{rel}^k$. The filters are fed into several 1 $\times$ 1 convolution layers, each of which uses ReLU as the activation function without normalization. Supposing $d'=16$, the output dimension of the intermediate layer is 8, and two convolution layers are used, the length of the generated filters are calculated as:

\begin{equation}
169 = \underbrace {(16+3) \times 8 + 8}_{conv1} + \underbrace{ 8 \times 1 + 1}_{conv2}
\label{eq:filter_calcu}
\end{equation}
The output is all convolutional filters (including weights and biases) flattened in a compact vector and can be predicted by an MLP layer.

\section{Optimal Transport Solution}
\begin{figure}
	\centering
	{
		\includegraphics[width=12cm]{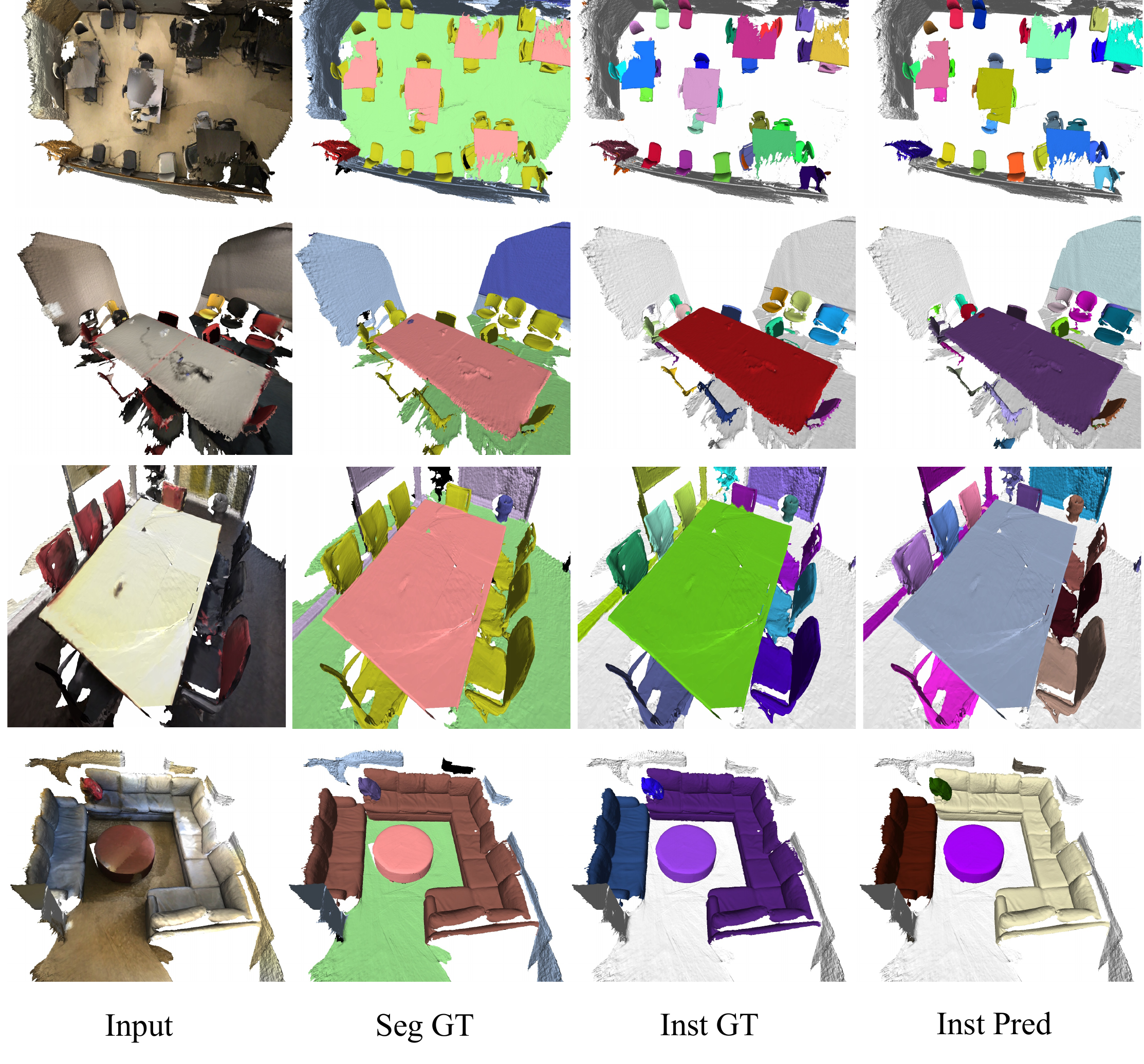}
	}
	
	\caption{Qualitative results of our method. 
	}
	
	\label{fig:res}
\end{figure}

In this section, we provide detailed descriptions of the Optimal Transport Solution for the dynamic targets assignment. The Optimal Transport problems are defined in terms of a set of suppliers and demanders, and the costs of transportation between them. We thus associated a demander with each prediction, and a supplier with each potential target. To address the negative samples, we add a background mask, filled with zero, to the target set. The goal is to optimally assign targets to samples. 
The algorithm is presented in Alg.~\ref{alg:pseudoOT} and only applied for training. 
In \textbf{Line1}, the network uses a sparseconv-based backbone and takes as input the point-wise coordinates \textbf{C} and features \textbf{F}. The output features of the backbone are denoted as ${F}_b=\{f_b^i\}_{i=1}^N$, where $N$ is the number of the input points. The mask features are denoted as $F_m$. 
In \textbf{Line2}, $K$ samples are selected from ${F}_b$ via the farthest sampling strategy, with features and coordinates denoted as $\{f_b^k\}_{k=1}^K$ and $\{p_b^k\}_{k=1}^K$, respectively.
In \textbf{Line3}, the selected samples are fed to the auxiliary instance head and $K$ masks $\{\hat{M^a_k}\}_{k=1}^K$ are predicted. The targets for supervising this head are consistent with the instance labels of the $K$ sampled points. For example, if the $k$-th point has an instance label of `$l_k$', the ground truth for the $k$-th mask is the binary mask representing the point set that has an identical instance label of `$l_k$'.
In \textbf{Line4-6}, the amount of supply for each foreground target is calculated based on the IoU between the foreground mask and the masks predicted by the auxiliary instance head.
In \textbf{Line7}, as each prediction requires one unit of the label (either instance or background), the total demands are $K$. To make sure that the total supply is equal to the total demands (see Eq. 2 in the main paper), we set the supply for the background target to be $K-\sum_{t=1}^T\mu_t$.
In \textbf{Line8}, we calculate the cost matrix according to Eq.3 (in the main paper).
In \textbf{Line9}, the demander vector is initialized with one, which has a length of $K$. This implies that the total demands for each prediction is one unit. 
In \textbf{Line10}, the optimal transportation plan is obtained by applying the Sinkhorn-Knopp algorithm \cite{Marco2013_sinkhorn}. Given $\bm U ^*$, the point will then be associated with the target that has allocated the greatest proportion of its supply. These recalibrated targets are applied for supervising the main instance head, which will be used to output the final predictions. More results are shown in Fig.\ref{fig:res}

\begin{algorithm}[t]
	\caption{Optimal Transport Solution}
	\label{alg:pseudoOT}
	\hspace*{0.02in} {\bf Input:} points with coordinates {C} and features {F}; \\
	\hspace*{0.44in} T masks for foreground instances $\{M_1,\dots M_T \}$ \\
	\hspace*{0.44in} $K$ is the number of randomly selected samples.\\
	\hspace*{0.44in} initialize a zero vector $\bm{\mu}_{T+1} $ with a length  of T+1 \\
	
	\hspace*{0.02in} {\bf Output:} Optimal Transport Plan $\bm U ^*$ \\
	
	\begin{algorithmic}[1]
		\State $\{f_b^i\}_{i=1}^N, {F}_m$ $\gets{ \text{Forward}({F}, {C})}$
		\State Randomly select $K$ samples: $\{f_b^k\}_{k=1}^K$, $\{p_b^k\}_{k=1}^K$
		\State $\{\hat{M_k^a} \}_{k=1}^K \gets \text{InstHEAD}_{\text{aux}}(\{f_k\}_{k=1}^K, \{p_k\}_{k=1}^K, F_m) $ 
		\For{t $\le$ T }
		\State $\mu_t$ = int($\sum_k${\text{IoU($M_t$, $\hat{M^a_k}$)}})
		\EndFor
		\State $\mu_{T+1}$ = K - $\sum_{t=1}^T{\mu_t}$
		\State Calculate cost matrix $\bm C$ according to Eq. 3
		\State $\bm {\nu_{K}} \gets \text{OnesInit}$
		\State $\bm U ^*$ = SinkHorn($\bm{\mu}_{T+1}$, $\bm C$, $\bm{\nu_{K}}$)

		\State \Return $\bm U ^*$
	\end{algorithmic}
\end{algorithm}

\bibliographystyle{splncs04}
\bibliography{egbib}
\end{document}